%% file: main.tex
\documentclass[letterpaper, 10 pt, conference]{ieeeconf}
\usepackage{times}
\usepackage{color, colortbl, xcolor}
\usepackage{multicol}
\usepackage[bookmarks=true]{hyperref}
\usepackage{amsfonts}
\usepackage{algorithmic}
\usepackage[ruled,vlined,titlenotnumbered]{algorithm2e} 
\usepackage{amsmath}
\usepackage{dsfont}
\usepackage{multicol}
\usepackage[pdftex]{graphicx}   
\usepackage{subcaption}

\usepackage{balance}

\usepackage[%
    style=numeric-comp,
    sorting=none,
    backend=biber,
    sortcites=true,
    doi=false,
    firstinits=true,
    hyperref,
    isbn=false,
    eprint=false,
    block=none]
    {biblatex}
    
    \renewbibmacro{in:}{}
    \AtEveryBibitem{%
  	\clearlist{language}%
  	\clearfield{pages}%
	}
\bibliography{references}

\input{notation}
\graphicspath{{./figures/}}    
\DeclareGraphicsExtensions{.pdf,.png}

\IEEEoverridecommandlockouts
\usepackage[textfont=md,font=footnotesize]{caption}

\usepackage{ifthen}
\newboolean{include-notes}
\setboolean{include-notes}{true}

\usepackage{lipsum}

\newcommand\blfootnote[1]{%
  \begingroup
  \renewcommand\thefootnote{}\footnote{#1}%
  \addtocounter{footnote}{-1}%
  \endgroup
}

\newcommand{\dfknote}[1]%
    {\textcolor{orange}{\textbf{DFK: #1}}}
\newcommand{\jfnote}[1]%
    {\textcolor{blue}{\textbf{JF: #1}}}
\newcommand{\remove}[1]%
    {\textcolor{red}{#1}}

\newcommand{\example}[1]%
{
\textbf{Running example:}
\textit{#1}
}

\begin{document}

\title{Safely Probabilistically Complete Real-Time Planning and Exploration\\in Unknown Environments}

\author{
David Fridovich-Keil$^*$, Jaime F. Fisac$^*$, and Claire J. Tomlin}


%

\maketitle

\blfootnote{
Department of Electrical Engineering and Computer Sciences,
UC Berkeley, \href{mailto:dfk@berkeley.edu}{\tt \small{\{dfk, jfisac, tomlin\}@berkeley.edu}}. This research is supported by an NSF CAREER award, the Air Force Office of Scientific Research (AFOSR), NSF's CPS FORCES and VeHICaL projects, the UC-Philippine-California Advanced Research Institute, the ONR MURI Embedded Humans, a DARPA Assured Autonomy grant, and the SRC CONIX Center.\\
$^*$The first two authors contributed equally to this work.}

\begin{abstract}
We present a new framework for motion planning that wraps around existing kinodynamic planners and guarantees recursive feasibility when operating in \emph{a priori} unknown, static environments. Our approach makes strong guarantees about overall safety and collision avoidance by utilizing a robust controller derived from reachability analysis. We ensure that motion plans never exit the safe backward reachable set of the initial state, while safely exploring the space. This preserves the safety of the initial state, and guarantees that that we will eventually find the goal if it is possible to do so while exploring safely. We implement our framework in the Robot Operating System (ROS) software environment and demonstrate it in a real-time simulation.
\end{abstract}

\IEEEpeerreviewmaketitle

\input{intro.tex}

\input{related_work.tex}

\input{problem.tex}

\input{framework.tex}

\input{example.tex}

\input{conclusion.tex}


\balance
\printbibliography

\end{document}

%% file: notation.tex
\newcommand{\env}{\mathcal{X}}
\newcommand{\fov}{\mathcal{F}}
\newcommand{\loc}{x}
\newcommand{\edim}{n_{\loc}}
\newcommand{\tsdim}{n_{\tstate}}
\newcommand{\psdim}{n_{\pstate}}
\newcommand{\tcdim}{n_{\tctrl}}
\newcommand{\pcdim}{n_{\pctrl}}
\newcommand{\ddim}{n_{\dstb}}
\newcommand{\rdim}{n_{\rstate}}
\newcommand{\ktrack}{k^*}
\newcommand{\frs}{\Omega_F}
\newcommand{\brs}{\Omega_B}

\newcommand{\teb}{\mathcal{E}}
\newcommand{\footprint}{\phi}
\newcommand{\envfree}{\env_\textrm{FREE}}
\newcommand{\home}{\pstate_{\textrm{home}}}
\newcommand{\goal}{\pstate_{\textrm{goal}}}
\newcommand{\explorable}{\pset_{\textrm{SE}}}
\newcommand{\sample}{\pstate_{\textrm{new}}}
\newcommand{\frg}{\mathcal{G}_F}
\newcommand{\brg}{\mathcal{G}_B}

\newcommand{\ptarget}{\mathcal{L}}

\newcommand{\zplanner}{z_{\textrm{p}}}

\newcommand{\pset}{\mathcal{P}} 
\newcommand{\tset}{\mathcal{S}} 

\newcommand{\tstate}{s} 
\newcommand{\pstate}{p} 
\newcommand{\rstate}{r} 
\newcommand{\ptraj}{\xi}

\newcommand{\tctrl}{u} 
\newcommand{\dstb}{d} 
\newcommand{\pctrl}{c} 
\newcommand{\tdyn}{f} 
\newcommand{\pdyn}{g} 
\newcommand{\dset}{\mathcal{D}}
\newcommand{\tcset}{\mathcal{U}}
\newcommand{\pcset}{\mathcal{C}}
\newcommand{\rset}{\mathcal{R}}

\newcommand{\bq}{\begin{equation}}
\newcommand{\eq}{\end{equation}}
\newcommand{\bmat}{\left[\begin{matrix}}
\newcommand{\emat}{\end{matrix}\right]}

\newcommand{\bqs}{\begin{equation*}}
\newcommand{\eqs}{\end{equation*}}

\def\atan2{\operatorname{atan2}}

\newtheorem{remark}{Remark}

\newtheorem{definition}{Definition}
\newtheorem{lemma}{Lemma}
\newtheorem{theorem}{Theorem}
\newtheorem{corollary}{Corollary}

%% file: intro.tex
\section{Introduction}
\label{sec:intro}

Motion planning is a foundational problem in mobile robotics, and the community has devoted significant effort to building theoretical and practical tools for a wide variety of applications.
Traditionally, the output of a motion planner is a desired plan or \emph{trajectory} for a dynamical system model.
This trajectory is then \emph{tracked} by one or more layers of low-level controllers.
Since the real, physical vehicle may follow higher-order, more complex dynamics than those used during planning, the trajectory it actually follows will not coincide with that which was planned.
This presents a problem for planners which aim to provide collision-avoidance guarantees.

\begin{figure}[htbp]
    \centering
    \includegraphics[width=\linewidth]{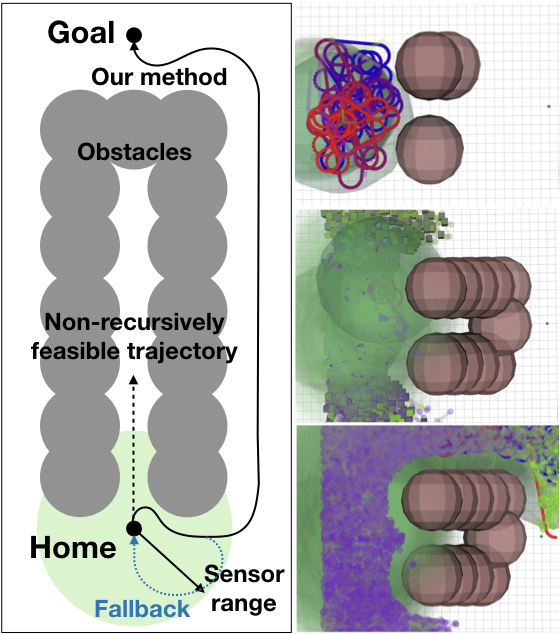}
    \caption{Depiction of our framework in operation, using a Dubins car model with a fixed minimum turning radius and constant speed.
    \emph{Left:} Schematic diagram of an environment in which a non-recursively feasible planning algorithm could enter a narrow dead end and fail to recover. \emph{Right:} Snapshots of our framework over time. 
    We build a search graph in known free space, identifying robustly viable trajectories that can safely return to the initial state or directly reach the goal. The physical system iteratively explores the environment along these recursively feasible plans and is eventually guaranteed to identify a viable trajectory to the goal, if one exists~(bottom~right).}
    \label{fig:front}
\end{figure}

Recently, the FaSTrack framework \cite{herbert2017fastrack} provides a mechanism for quantifying the maximum \textit{tracking error} between a high-order dynamical model of the physical system and a lower-order model used for planning. This analysis can be done offline, using a reachability computation, and supplied to a real-time motion planner for online collision-checking. Other, similarly motivated work, e.g. \cite{Singh2017}, also seeks to quantify this maximum tracking error.

Still, a key challenge remains: in \emph{a priori} unknown environments where obstacles are sensed online, it can be difficult to guarantee \emph{recursive feasibility}. Informally, a planning algorithm is recursively feasible if it explores the environment safely and without losing its ability to reach the goal. The dangers of unsafe exploration are illustrated in Fig.~\ref{fig:front} (left), in which a non-recursively feasible planner enters a dead end which the system cannot exit. Most motion planners bypass these issues, for example by assuming full prior knowledge of the environment or by assuming that it is safe to stop and possible to do so instantaneously. While such techniques are effective in many scenarios, there are important applications and systems for which safe exploration really does matter, e.g. a fixed-wing aircraft operating with limited visibility. More generally, it is important to consider recursive feasibility for systems such as unicycles, bicycles, and cars that have inertia and function at relatively high speeds. These issues are especially pronounced for non-holonomic systems.

We propose building a graph of forward-reachable states (for a given dynamical system planning model) within known free space, while simultaneously identifying those states from which the initial state is reachable. This graph implicitly represents a discrete under-approximation of the backward-reachable set of the initial state.
We guarantee the safety of the physical system, modeled with higher-order dynamics, using a robust control scheme \cite{herbert2017fastrack}.
Our framework, illustrated in Fig.~\ref{fig:front}, ensures:
\begin{itemize}
\item \emph{Safety:} all trajectories initiated by the physical system will be robustly collision-free.
\item \emph{Liveness:} if the goal is safely reachable from the initial state, it will always be safely reachable. 
\item \emph{Safe Probabilistic Completeness:} if a goal was originally reachable by a plan that preserves the ability to return home, our framework guarantees that it will eventually be found with probability 1.
\end{itemize}

%% file: related_work.tex
\section{Related Work}
\label{sec:related_work}

Though we defer a formal definition until Section~\ref{subsec:definitions}, ultimately, a motion planner is recursively feasible if it can explore unknown space while always remaining safe. There is an extensive body of literature in motion planning and safe exploration, which we cannot hope to fully summarize here.
Rather, we identify two main categories of related work and discuss several of the most closely related approaches.

\subsection{Safe motion planning}
\label{subsec:safe_motion_planning}

Recent methods such as \cite{herbert2017fastrack, Majumdar2017, Singh2017, fridovich2018planning} provide a variety of mechanisms for robust motion planning. Here, robustness is characterized in terms of an envelope around planned trajectories which the physical system is guaranteed to remain within. Our work relies upon this idea, building upon~\cite{herbert2017fastrack}.

However, robust planning does not automatically guarantee recursive feasibility. Richards and How~\cite{richards2003model} and Rosolia and Borrelli~\cite{rosolia2018learning} directly address this problem within a model predictive control framework. The major differences between these works and our own are that~\cite{richards2003model} assumes linear time-invariant system dynamics, while~\cite{rosolia2018learning} addresses an iterative task. Moreover, both assume \emph{a priori} knowledge of all obstacles. Schouwenaars et. al.~\cite{schouwenaars2004decentralized} also plan in a receding horizon, but as in our work, recursive safety (though not liveness) is guaranteed by ensuring that all planned trajectories terminate in a safe loiter pattern.

Our work may also be viewed as an extension of graph-based kinodynamic planners, e.g. the probabilistic roadmap~\cite{svestka1996probabilistic}, by enforcing that all edges in the graph are part of safely executable trajectories. Importantly, our framework guarantees recursive feasibility in an \emph{a priori} unknown environment, with potentially high-order system dynamics, and in the presence of environmental disturbances.

\subsection{Safe exploration} 
\label{subsec:safe_exploration}

There is a rich body of work in robotic exploration methods, which tackle the problem of finding viable trajectories to a specified goal in an initially unknown environment. The majority of proposed methods, such as frontier-exploration \cite{yamauchi1997frontier, yamauchi1998frontier, yoder2016autonomous} and D* \cite{stentz1994optimal, koenig2005fast}, have traditionally operated in configuration space, assuming a \emph{kinematic} model of the robot's motion. Our method, in contrast, focuses on robotic systems for which a \emph{dynamic} model is necessary, such as autonomous cars and aircraft. Bekris and Kavraki \cite{bekris2007greedy} present a sampling-based strategy which reasons about inevitable collision sets \cite{fraichard2004inevitable}, but is restricted to work with drift-less dynamics. More recent work by Janson et. al. \cite{janson2018safe} also addresses the dynamic exploration problem, but assumes that the vehicle is able to come to a stop in finite time. 

Exploration has also been studied within the context of Markov Decision Processes (MDPs) and Reinforcement Learning (RL).
Moldovan and Abbeel \cite{moldovan2012safe} propose an approach for generating a sequence of actions which preserve ergodicity with high probability. 
Other similar approaches, e.g. \cite{achiam2017constrained}, also design risk-aware control policies that satisfy approximate constraints. 
Berkenkamp et. al. \cite{berkenkamp2017safe} and Chow et. al. \cite{chow2018lyapunov} define safety in terms of Lyapunov stability. Though generally desirable, stability is insufficient to guarantee collision avoidance; in this work, we use a stronger set-based definition of safety. Our formulation of safe exploration is closely related with that of \cite{akametalu2014reachability,fisac2018general, fraichard2004inevitable}, which characterize safety with reachable sets.


%% file: problem.tex
\section{Problem Formulation}
\label{sec:problem}

\subsection{Preliminaries}
\label{subsec:preliminaries}

We consider an autonomous navigation task in a bounded \emph{a priori} unknown static environment  $\env\subset\mathbb{R}^{\edim}$.
The autonomous system has dynamic state $\tstate\in\tset\subset\mathbb{R}^{\tsdim}$, which includes, but is in general not limited to its location $\loc$ in the environment $\env$.
We presume that for each point $\loc \in \env$, the environment representation can assign a label $\{\textrm{OCCUPIED}, \textrm{FREE}, \textrm{UNKNOWN}\}$.
The system's knowledge of the environment will be updated online according to measurements from a well-characterized sensor, with field of view $\fov:\tset\to 2^{\env}$.
In this work, we will restrict our attention to deterministic sensing models, i.e. if $x \in \env$ is within the sensor's field of view $\fov(s)$, it will be correctly identified as either $\{\textrm{OCCUPIED}, \textrm{FREE}\}$.
Probabilistic extensions are possible, though beyond the scope of this paper.

We assume known system dynamics, of the form:
\begin{align*}
    \dot\tstate = \tdyn(\tstate,\tctrl,\dstb)
    \enspace,
\end{align*}
where $\tctrl\in\tcset\subset\mathbb{R}^{\tcdim}$ is the system's control input
and $\dstb\in\dset\subset\mathbb{R}^{\ddim}$ is a bounded disturbance.

In general, the dynamical model $\tdyn$ of the physical system will be nonlinear and high-order,
making it challenging to compute trajectories in real time.
Instead, we can use an approximate, lower-order dynamical model for real-time trajectory computation, along with a framework which produces a known tracking controller for the full-order model allowing it to follow the trajectories of the low-order model with a guarantee on accuracy.
Let the simplified state of the system for planning purposes be $\pstate\in\pset\subset\mathbb{R}^{\psdim}$, governed by approximate \emph{planning dynamics}:
\begin{align*}
    \dot\pstate  = \pdyn(\pstate,\pctrl)
    \enspace,
\end{align*}
with $\pctrl\in\pcset\subset\mathbb{R}^{\pcdim}$ the control input of the simplified system, which we will refer to as the \emph{planning system}.

We use the FaSTrack framework~\cite{herbert2017fastrack} to provide a robust controlled invariant set in the relative state space $\rset\subset\mathbb{R}^{\rdim}$ between the planning reference and the full system.
This relative state depends on the dynamical models used.
A concrete example will be presented in Section \ref{sec:example}, and
we direct the reader to~\cite{herbert2017fastrack,fridovich2018planning} for further discussion.
The output of this robust analysis is two-fold:
the autonomous system is given an optimal tracking control law $\ktrack:\rset\to\tcset$ that will keep the relative state inside of this invariant set at runtime \emph{regardless} of the low-order trajectory proposed by the planning algorithm.
In turn, the planning algorithm can use the projection of the invariant set onto the planning state space $\pset$ as a guaranteed tracking error bound $\teb\subset\pset$ for the purposes of collision-checking at planning time.
A feature of the FaSTrack framework is that the robust safety analysis depends only upon the relative dynamics, and \emph{not} on the particular algorithm used for planning low-order trajectories.
We inherit this modularity in our recursively feasible planning framework, which can be used with an arbitrary low-level motion planner.
In Section \ref{sec:example}, we demonstrate our framework with a standard third-party algorithm from the Open Motion Planning Library (OMPL)~\cite{sucan2012ompl}.


\subsection{Recursive Feasibility: Safety and Liveness}
\label{subsec:definitions}

We now define several important concepts more formally, as they pertain directly to the theoretical safety guarantees of our proposed framework.
Let ${\ptraj\big(\cdot;t_0,\pstate,\pctrl(\cdot)\big):\mathbb{R}\to\pset}$ denote the trajectory followed by the planning system starting at state $\pstate$ at time $t_0$ under some control signal $\pctrl(\cdot)$ over time.

Given a planned state $\pstate$, we refer to its \emph{footprint} $\footprint(\pstate)$ as the set of points $\loc\in\env$ that are occupied by the system in this state.
We additionally define the robust footprint $\footprint_{\teb}(\pstate)$ as the set of points $\loc\in\env$ that are occupied by some $\pstate'\in\{p+\teb\}$ (with $+$ here denoting Minkowski addition).
This represents the set of locations that may be occupied by the physical system while attempting to track the planned state $\pstate$.
We will require that the system is at all times guaranteed to only occupy locations known to be $\textrm{FREE}$. For convenience, we will denote by $\envfree(t)$ the set of points $\loc\in\env$ that are labelled as $\textrm{FREE}$ at time $t$.


We then have the following definitions.
\begin{definition}{(Safety)}
\label{def:safety}
A planned trajectory $\ptraj\big(\cdot; t_0, \pstate,\pctrl(\cdot)\big)$ is known at time $t_0$ to be safe, i.e. collision-free, if it satisfies the following criterion:
\begin{align*}
    \forall t \ge t_0, \footprint_{\teb}\Big(\ptraj\big(t;t_0, \pstate, \pctrl(\cdot)\big)\Big) \subseteq \envfree(t_0)
    \enspace.
\end{align*}
\end{definition}

Observe that Definition \ref{def:safety} is \emph{not} a statement about stability, as in e.g. \cite{berkenkamp2017safe}.
Dynamic stability is in fact neither a necessary nor a sufficient condition for safety understood as guaranteed collision (and failure) avoidance.

\begin{definition}{(Safe Reachable Set)}
\label{def:safe_reachable_set}
The safe forward reachable set $\frs$ of a set of states $\ptarget\subseteq\pset$ at time $t_0$ is the set of states $\pstate'\in\pset$ that are known at $t_0$ to be safely reachable from $\ptarget$ under some control signal $\pctrl(\cdot)$.
\begin{align*}
    &\frs(\ptarget;t_0) := \Big\{\pstate' \mid \exists \pstate\in\ptarget, \exists t \ge t_0, \exists \pctrl(\cdot), \forall \tau\in[t_0,t] :\\
    &~\footprint_{\teb}\Big(\ptraj\big(\tau; t_0,\pstate, \pctrl(\cdot)\big)\Big)\subseteq\envfree(t_0), \pstate' = \ptraj\big(t; t_0,\pstate, \pctrl(\cdot)\big)\Big\}
    \,.
\end{align*}
Analogously, the safe backward reachable set $\brs$ of $\ptarget$ at $t_0$ is the set of states $\pstate'\in\pset$ from which $\ptarget$ is known at time $t_0$ to be safely reachable under some control signal
(this can also be thought of as the set of states $\pstate'\in\pset$ that can be safely reached from $\ptarget$ in backward time, hence the name \emph{backward} reachable set):
\begin{align*}
    &\brs(\ptarget;t_0) := \Big\{\pstate' \mid \exists \pstate\in\ptarget, \exists t \ge t_0, \exists \pctrl(\cdot), \forall \tau\in[t_0,t] :\\
    &~\footprint_{\teb}\Big(\ptraj\big(\tau; t_0,\pstate', \pctrl(\cdot)\big)\Big)\subseteq\envfree(t_0), \pstate = \ptraj\big(t; t_0,\pstate', \pctrl(\cdot)\big)\Big\}
    \,.
\end{align*}
We will often consider reachable sets of individual states; for conciseness, we will write $\brs(\pstate; t_0)$ rather than $\brs(\{\pstate\}, t_0)$.

\end{definition}

We now proceed to define viability in terms of these sets.

\begin{definition}{(Viability)}
\label{def:viability}
    A state $\pstate$ is \emph{viable} at time $t_0$ with respect to a goal state $\goal$ and a home state $\home$ if at $t_0$ it is known to be possible to reach either $\goal$ or $\home$ from $\pstate$ while remaining safe, i.e. ${\pstate\in\brs\big(\{\goal, \home\}; t_0\big)}$.
    A trajectory $\ptraj$ is viable at $t_0$ if all states along $\ptraj$ are viable at $t_0$. Note that a trajectory can be safe (Def.~\ref{def:safety}) but not viable.
\end{definition}

\begin{definition}{(Safely Explorable Set)}
\label{def:safely_explorable_set}
    The safely explorable set $\explorable(\pstate)\subset\pset$ of a state $\pstate$ is the collection of states that can eventually be visited by the system through a trajectory starting at state $\pstate$ with no prior knowledge of $\env$ whose states are,
    at each time $t \ge 0$,
    viable according to the known free space $\envfree(t)$.
\end{definition}

Based on the idea of the safely explorable set we can now introduce the important notion of \emph{liveness} for the purposes of our work.
\begin{definition}{(Liveness)}
\label{def:liveness}
    A state $\pstate$ is live with respect to a goal state $\goal$ if it is possible to reach $\goal$ from $\pstate$ while remaining in the safely explorable set for all time, i.e if ${\goal \in \explorable(\pstate)}$.
    A trajectory $\ptraj$ is live if all states in $\ptraj$ are live.
\end{definition}

Finally, we will refer to a planning algorithm as \emph{recursively feasible} if, given that the initial state $\pstate_0$ is live, all future states $\pstate$ are both live and viable. We will show that our proposed framework is recursively feasible. Moreover, we will also show that it is \textit{safely probabilistically complete}, in the sense that, if $\pstate_0$ is live with respect to $\goal$, then we will eventually reach $\goal$ through continued guaranteed safe exploration, with probability 1.


%% file: framework.tex
\section{General Framework}
\label{sec:framework}

\subsection{Overview}
\label{subsec:overview}

Our framework is comprised of two concurrent, asynchronous operations: building a graph of states which discretely under-approximate the forward and backward reachable sets of the initial ``home'' state, and traversing this graph to find recursively feasible trajectories. Namely, we define the graph $\frg := \{V, E\}$ of vertices $V$ and edges $E$. Vertices are individual states in $\pset$, and directed edges are trajectories $\ptraj$ between pairs of vertices. $\frg$ will be a discrete under-approximation of the current safe forward reachable set of the initial state $\home$. We also define the graph $\brg \subseteq \frg$ to contain only those vertices which are in the safe backward reachable set of $\home$ and $\goal$, and the corresponding edges. We use the notation $\pstate \in \frg$ to mean that state $\pstate$ is a vertex in $\frg$, and likewise for $\brg$.

We use following two facts extensively. They follow directly from the definitions above and our assumptions on deterministic sensing and a static environment. 
\begin{remark}{(Permanence of Safety)}
\label{rem:permanence_safety}
    A trajectory $\ptraj$ that is safe at time $t_0$ will continue to be safe for all ${t\ge t_0}$.
\end{remark}

\begin{remark}{(Permanence of Reachability)}
\label{rem:permanence_reachability}
    A state $\pstate$ that is in the safe forward or backward reachable set of another state $\pstate_0$ at time $t_0$ will continue to belong to this set for all $t\ge t_0$, i.e. $\frs(\pstate_0; t_0) \subseteq \frs(\pstate_0; t)$ and $\brs(\pstate_0; t_0) \subseteq \brs(\pstate_0; t)$.
\end{remark}


\subsection{Building the graph}
\label{subsec:building_graph}

We incrementally build the graph by alternating between outbound expansion and inbound consolidation steps.
In the outbound expansion step, new candidate states are sampled, and if possible, connected to $\frg$. This marks them as part of the forward reachable set of $\home$.
In the inbound consolidation step, we attempt to find a safe trajectory from forward-reachable states in $\frg$ back to a state in $\brg$, which is known to be \emph{viable}.
Successful inbound consolidation marks a state as either able to reach $\goal$ or safely return to $\home$.

\subsubsection{Outbound expansion}
\label{subsubsec:outbound}

This process incrementally expands a discrete under-approximation $\frg$ of the forward reachable set of the home state, $\frs(\home;t)$.
Note that, by Remark~\ref{rem:permanence_reachability}, $\frs(\home;t)$ can only grow as the environment $\env$ is gradually explored over time and therefore any state $\pstate$ added to  $\frg$ at a given time $t$ is guaranteed to belong to $\frs(\home;t')$ for all $t'\ge t$.

\begin{figure}
\centering
\begin{subfigure}[t]{0.47\columnwidth}
    \centering
    \includegraphics[width=\linewidth]{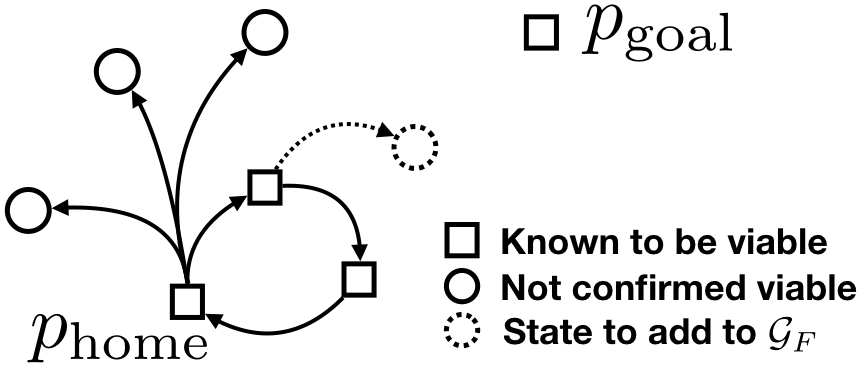}
    \caption{Outbound expansion.}
    \label{fig:outbound_expansion}
\end{subfigure}
~
\begin{subfigure}[t]{0.47\columnwidth}
    \centering
    \includegraphics[width=\linewidth]{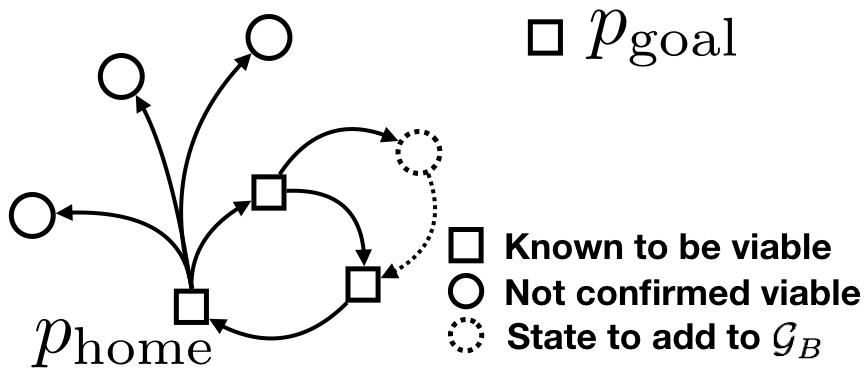}
    \caption{Inbound consolidation.}
    \label{fig:inbound_consolidation}
\end{subfigure}
\caption{In outbound expansion (a), a new state is sampled from $\pset$ and added to $\frg$ if safely reachable from $\frg$. In inbound consolidation (b) a state in $\frg$ is added to $\brg$ if it can safely reach a (viable) state in $\brg$.}
\label{fig:graph_building}
\end{figure}

We add states to $\frg$ via a Monte Carlo sampling strategy inspired by existing graph-based kinodynamic planners \cite{karaman2011RRTPRM}, illustrated in Fig.~\ref{fig:outbound_expansion}.
We present a relatively simple strategy here, although more sophisticated options for sampling new states are possible, e.g. \cite{karaman2013sampling,gammell2015batch}.

Let $\sample$ be sampled uniformly at random from $\pset$ at time $t$ such that $\footprint_{\teb}(\sample) \subseteq \envfree(t)$.
We wish to establish whether or not $\sample$ is in the safe forward reachable set of home at $t$, i.e. $\sample \in \frs(\home; t)$.
This is accomplished by invoking a third-party motion planner, which will attempt to find a safe trajectory to $\sample$ from any of the points already known to be in $\frs(\home; t)$. In Section~\ref{sec:example}, we use a standard kinodynamic planner from the OMPL \cite{sucan2012ompl} for this purpose.

We observe that repeatedly executing this procedure will, in the limit, result in a dense discrete under-approximation of $\frs(\home;t)$.
Formally, assuming that the low-level planner will find a valid trajectory to a sampled state $\pstate$ if one exists, then for any $\epsilon>0$, we have that the probability that a new sampled state $\pstate'\in\frs(\home,t)$ will lie within a distance of $\epsilon$ from the nearest state $\pstate\in\frg$ goes to $1$ in the limit of infinite samples. We formalize this observation below:
\begin{lemma}{(Dense Sampling)}
\label{lem:dense_sampling}
For all $\epsilon > 0$, assuming we sample candidate states $\pstate$ uniformly and independently from $\pset$ and $\pset$ is compact, then letting $\pstate_k$ be the $k$-th sampled state from $\pset$ we have that $\forall t$:
\begin{align*}
    \lim_{k\to\infty}P\big( \min_{\pstate\in\frg}\|\pstate_k - \pstate\|<\epsilon\mid \pstate_k\in\frs(\home;t)\big) = 1
    \enspace.
\end{align*}
\begin{proof}
    This follows directly from the properties of uniform sampling from compact sets.
\end{proof}
\end{lemma}
This will be useful in proving the safe probabilistic completeness of our framework.

\subsubsection{Inbound consolidation}
\label{subsubsec:inbound}

This process incrementally adds states in $\frg$ to a discrete approximation $\brg$ of the safe \emph{backward} reachable set of $\{\home, \goal\}$.
By Definition~\ref{def:viability}, any state added to this set is \emph{viable},
which means that a trajectory will always exist from it to either $\goal$ or $\home$.
This is a crucial element of our overall guarantee of recursive feasibility.
We recall that $\brg \subseteq \frg$.

Suppose that $\pstate \in \frg \setminus \brg$. We will attempt to add $\pstate$ to $\brg$ by finding a safe trajectory from $\pstate$ to any of the states currently in $\brg$ by invoking the low-level motion planner.
If we succeed in finding such a trajectory, then by construction there exists a trajectory all the way to $\home$,
so we add $\pstate$ to $\brg$. 
If $\pstate$ is added to $\brg$, we also add all of its ancestors in $\frg$ to $\brg$, since there now exists a trajectory from each ancestor through $\pstate$ to either $\home$ or $\goal$. This procedure is illustrated in Fig.~\ref{fig:inbound_consolidation}.



\subsection{Exploring the graph}
\label{subsec:exploring_graph}

When requested, we must be able to supply a safe trajectory beginning at the current state reference $\pstate(t)$ tracked by the system.
Recall from Section~\ref{subsec:preliminaries} that under the robust tracking framework~\cite{herbert2017fastrack}, the physical system's state $\tstate(t)$ is guaranteed to remain within an error bound $\teb$ of $\pstate(t)$ measured on the planning state space $\pset$. This property allows us to make guarantees in terms of planning model states $\pstate$ rather than full physical system states $\tstate$.

Trajectories~$\ptraj$ output by our framework must guarantee future safety for all time;
that is, as the system follows~$\ptraj$ we must always be able to find a safe trajectory starting from any future state.
In addition, we require that $\home$ remains safely reachable throughout the trajectory;
this ensures that \emph{liveness} is preserved (if it was possible from $\home$ to safely explore $\env$ and reach $\goal$ then this possibility will not be lost by embarking on~$\ptraj$).
Note that liveness is an important property separate from safety:
a merely safe planner may eventually trap the system in a periodic safe orbit that it cannot safely exit.

By construction, any cycle in $\brg$ is safe for all future times (Remark~\ref{rem:permanence_safety}).
Readily, this suggests that we could guarantee perpetual recursive feasibility by always returning the same cycle.
However, this naive strategy would never reach the goal.
Moreover, it would not incrementally explore the environment.
In order to force the system to explore unknown regions of $\env$, we modify this naive strategy by routing the system
through a randomly selected \emph{unvisited} state $\sample\in\brg$, and then back to $\home$.
The trajectory always ends in a periodic safe orbit between $\sample$ and $\home$.
Note that this random selection does not need to be done naively (e.g. by uniform sampling of unvisited states in $\brg$), and efficient exploration strategies are certainly possible.
In our examples we will use an $\epsilon$-greedy sampling heuristic by which,
with probability $1-\epsilon$, we select the unvisited $\pstate\in\brg$ closest to $\goal$,
and otherwise, with probability $\epsilon$, we uniformly sample an unvisited state in $\brg$.


Of course, if $\goal$ is ever added to $\brg$, we may simply return a trajectory from the current state $\pstate(t)$ to $\goal$.
This will always be possible because, by construction, every state in $\brg$ is safely reachable from every other state in $\brg$ (if necessary, looping through $\home$).


\subsection{Algorithm summary}
\label{subsec:algorithm_summary}

To summarize, our framework maintains graph representations of the forward reachable set of $\home$ and the backward reachable set of $\{\home,\goal\}$. Over time, these graphs become increasingly dense (Lemma~\ref{lem:dense_sampling}). Additionally, all output trajectories terminate at $\goal$ or in a cycle that includes $\home$. This implies our main theoretical result:

\begin{theorem}{(Recursive Feasibility)}
\label{thm:recursive_feasibility}
Assuming that we are able to generate an initial viable trajectory (e.g. a loop through $\home$), all subsequently generated trajectories will be viable and preserve the liveness of $\home$.
Thus, our framework guarantees \emph{recursive feasibility}.\\
\begin{proof}
    By assumption, the initial trajectory $\ptraj_0$ output at $t_0$ is safe (Definition~\ref{def:safety}).
    We now proceed by induction: assume that the $i$-th reference trajectory $\ptraj_i$ is viable for the knowledge of free space at the time $t_i$ at which it was generated, i.e. $\forall t\ge t_i, \ptraj_i(t)\in\brs(\{\home,\goal\};t_i)$.
    Assuming $\goal$ has not been reached yet at the time of the next planning request, $t_{i+1}$, a new trajectory will be generated from initial state $\ptraj_i(t_{i+1})$.
    The new trajectory $\ptraj_{i+1}$ will be created by concatenating safe trajectories between states in $\brg\subseteq\brs(\{\home,\goal\};t_i)$ and therefore will be a viable trajectory.
    Such a trajectory can always be found, because it is always possible to choose $\ptraj_{i+1}\equiv\ptraj_i$, which, by the inductive hypothesis was a viable trajectory at time $t_i$ and, by Remark~\ref{rem:permanence_reachability}, continues to be viable at $t_{i+1}$.
    Therefore all planned trajectories $\ptraj_i$ will retain the ability to either safely reach $\goal$ or safely return to $\home$. In the former case, $\ptraj_i$ is immediately live (and since $\forall t\ge 0, \ptraj_i(t)\in\frs(\home; t_i)$, $\home$ must have been live too); in the latter, $\ptraj_i$ will inherit the liveness of $\home$, by observing that $\forall t\ge 0, \ptraj_i(t)\in\brs(\home;t_i)$.
\end{proof}
\end{theorem}

\begin{corollary}(Dynamical System Exploration)\label{cor:dynamical_system}
    Given that the safety of trajectories is evaluated using the robust footprint $\footprint_\teb(\cdot)$, and the relative state between the dynamical system and the planning reference is guaranteed to be contained in $\teb$, Theorem~\ref{thm:recursive_feasibility} implies that the dynamical system can continually execute safe trajectories in the environment.
\end{corollary}

Moreover, we ensure that each output trajectory visits an unexplored state in $\brg$, which implies that $\brg$ approaches the safely explorable set $\explorable(\home)$ from Definition~\ref{def:safely_explorable_set}. Together with Theorem~\ref{thm:recursive_feasibility}, this implies the following completeness result:

\begin{theorem}{(Safe Probabilistic Completeness)}
\label{thm:completeness}
In the limit of infinite runtime, our framework eventually finds the goal
with probability $1$
if it is within the safely explorable set.\\
\begin{proof}
    By Theorem~\ref{thm:recursive_feasibility}, all trajectories output will be viable; hence, the autonomous system will remain safe for all time (Corollary~\ref{cor:dynamical_system}).
    Further, since each generated trajectory visits a previously unvisited state in $\brg$ with nonzero probability, by Lemma~\ref{lem:dense_sampling} it will eventually observe new regions in the safely explorable set $\explorable(\home)$ if any exist. Moreover, those regions will eventually be sampled, added to $\brg$, and visited by subsequent trajectories. Because we have assumed all sets of interest to be bounded, this implies that we will eventually add $\goal$ to $\brg$ as long as $\goal\in\explorable(\home)$.
\end{proof}
\end{theorem}


\subsection{Remarks}
\label{subsec:remarks}

We conclude this section with several brief remarks regarding implementation. 

In Sec.~\ref{subsec:building_graph}, we specify that states should be connected to existing states in $\frg$ and $\brg$. In practice, we find that connecting to one of the $k$-nearest neighbors (measured in the Euclidean norm over $\pset$) in the appropriate graph suffices. 

In Sec.~\ref{subsec:exploring_graph}, we describe traversing $\brg$ to find safe trajectories between vertices. For efficiency, we recommend maintaining the following at each vertex: cost-from-home, cost-to-home, and cost-to-goal, where cost may be any consistent metric on trajectories (e.g. duration). If these quantities are maintained, then care must be taken to update them appropriately for descendants and ancestors of states that are added to $\frg$ and $\brg$ in Sec.~\ref{subsec:building_graph}.

Finally, we observe that outbound expansion, inbound consolidation, and graph exploration may all be performed in parallel and asynchronously.


%% file: example.tex
\section{Example}
\label{sec:example}

We demonstrate our framework in a real-time simulation, implemented within the Robot Operating System (ROS) software environment \cite{quigley2009ros}.

\subsection{Setup}
\label{subsec:setup}

Let the high-order system dynamics be given by the following 6D model:
\begin{align}
\label{eqn:high_order_dynamics}
\dot \tstate =
\begin{bmatrix}
\dot x \\
\dot v_x \\
\dot y \\
\dot v_y \\
\dot z \\
\dot v_z
\end{bmatrix} = 
\begin{bmatrix}
v_x  \\
g \cos \tctrl_1 \\
v_y \\
-g \sin \tctrl_2 \\
v_z \\
\tctrl_3 - g
\end{bmatrix}
\end{align}
\noindent where $g$ is acceleration due to gravity, the states are position and velocity in $(x, y, z)$, and the controls are $\tctrl_1 = \textrm{pitch}, \tctrl_2 = \textrm{roll},~\textrm{and}~\tctrl_3 = \textrm{thrust acceleration}$. These dynamics are a reasonably accurate model for a lightweight quadrotor operating near a hover and at zero yaw.

We consider the following lower-order 3D dynamical model for planning:
\begin{align}
\label{eqn:low_order_dynamics}
\dot \pstate =
\begin{bmatrix}
\dot x \\
\dot y \\
\dot \theta
\end{bmatrix} = 
\begin{bmatrix}
v \cos \theta  \\
v \sin \theta \\
\pctrl
\end{bmatrix}
\end{align}
\noindent where $v$ is a constant tangential speed in the Frenet frame, states are absolute heading $\theta$, and $(x, y)$ position in fixed frame, and control $\pctrl$ is the turning rate. We interpret these dynamics as a Dubins car operating at a fixed $z$ height $\zplanner$.

\begin{figure}
\centering
\begin{subfigure}[t]{0.35\columnwidth}
    \centering    \includegraphics[width=\linewidth]{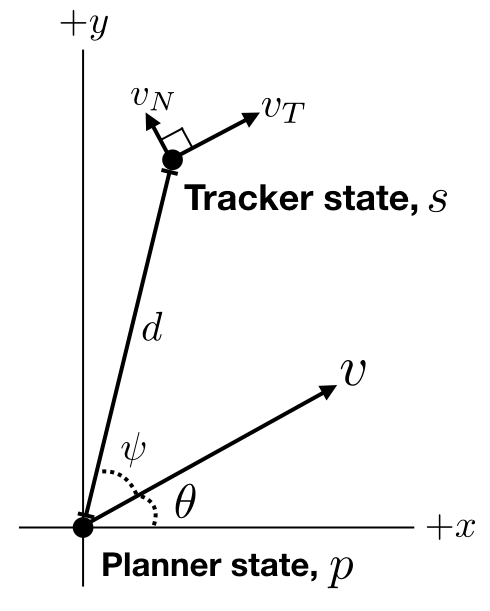}
    \caption{Relative states.}
    \label{fig:relative_state}
\end{subfigure}
~
\begin{subfigure}[t]{0.55\columnwidth}
    \centering    \includegraphics[width=\linewidth]{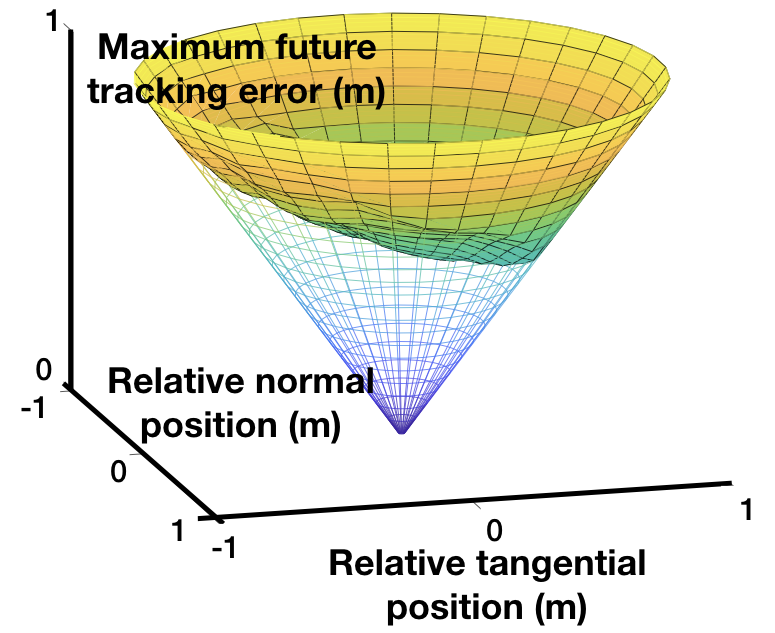}
    \caption{Computed value function.}
    \label{fig:value_function}
\end{subfigure}
\caption{(a) Relative states for 6D near-hover quadrotor tracking 3D Dubins car. (b) Minimum value over $v_T$ and $v_N$, for each relative $(x, y)$ position in the planner's Frenet frame. Any non-empty sublevel set can be used as a tracking error bound $\teb$.}
\label{fig:fastrack_precomputation}
\end{figure}

We take controls to be bounded in all dimensions independently by known constants: $\tctrl \in [\underline{\tctrl}_1, \overline{\tctrl}_1] \times [\underline{\tctrl}_2, \overline{\tctrl}_2] \times [\underline{\tctrl}_3, \overline{\tctrl}_3]$ and $\pctrl \in [\underline{\pctrl}, \overline{\pctrl}]$. 
In order to compute the FaSTrack tracking error bound $\teb$, we must solve a Hamilton Jacobi (HJ) reachability problem for the \emph{relative dynamics} defined by \eqref{eqn:high_order_dynamics} and \eqref{eqn:low_order_dynamics}. In this case, the relative dynamics are given by:
\begin{align}
\label{eqn:relative_dynamics}
\dot \rstate = 
\begin{bmatrix}
\dot d \\
\dot \psi \\
\dot v_T \\
\dot v_N
\end{bmatrix} = 
\begin{bmatrix}
v_T \cos \psi + v_N \sin \psi \\
-\pctrl - v_T \sin \psi + v_N \cos \psi \\
\tctrl_1 \cos \theta - \tctrl_2 \sin \theta + \pctrl v_T \\
-\tctrl_1 \sin \theta - \tctrl_2 \cos \theta - \pctrl v_T
\end{bmatrix}
\end{align}
\noindent with the relative states $d$ (distance), $\psi$ (bearing), $v_T$ (tangential velocity), and $v_N$ (normal velocity) illustrated in Fig.~\ref{fig:relative_state}.

Fig.~\ref{fig:value_function} is a 3D rendering of the FaSTrack value function \cite{herbert2017fastrack} computed using level set methods \cite{LSToolbox}. The value function records the maximum relative distance between the high- and low-order dynamical models (i.e. $d$). In order to guarantee that, at run-time, the distance between the two systems does not exceed this value, the value function is computed by solving a differential game in which $\pctrl(\cdot)$ seeks to maximize the relative distance and $\tctrl(\cdot)$ seeks to minimize it.
Observe in Fig.~\ref{fig:value_function} that level sets of the value function with sufficiently high value are well-approximated by discs centered on the origin in $(x, y)$. Thus, we approximate the TEB $\teb$ by such a disc for rapid collision-checking during each call to the low-level motion planner. Since the high-order dynamics \eqref{eqn:high_order_dynamics} do allow for variation in $z$, we also incorporate a $z$ extent for $\teb$ which may be obtained by solving a similar differential game in the $(z, v_z)$ subsystem of \eqref{eqn:high_order_dynamics}, as in \cite{fridovich2018planning}.

We use the KPIECE1 kinodynamic planner \cite{csucan2009kinodynamic} within the Open Motion Planning Library (OMPL) \cite{sucan2012ompl} to plan all trajectories for the low-level dynamics while building the graphs $\frg$ and $\brg$. For simplicity, we model static obstacles as spheres in $\mathbb{R}^3$ and use an omnidirectional sensing model in which all obstacles within a fixed range of the vehicle are sensed exactly. We emphasize that these choices of environment and sensing models are deliberately simplified in order to more clearly showcase our framework. The framework itself is fully compatible with arbitrary representations of static obstacles and deterministic sensing models. Extensions to dynamic obstacles and probabilistic sensing are promising directions for future research.


\subsection{Simulation Results}
\label{subsec:simulation_results}

We demonstrate our framework in a simple simulated environment, shown in Fig.~\ref{fig:front}, designed to illustrate the importance of maintaining recursive feasibility.
This simulation is intended as a proof of concept; our central contribution is theoretical and applies to a range of planning problems.\footnote{Video summary available at: \url{https://youtu.be/GKQwFxdJWSA}}

Observe in Fig.~\ref{fig:front} that our method avoids collision where a non-recursively-feasible approach would likely fail. Here, the goal is directly in front of the home position and the way there \emph{appears} to be in $\envfree(t)$. However, just beyond our sensor's field of view $\fov$, there is a narrow dead end. Many standard planning techniques would either optimistically assume the unknown regions of the environment are free space, or plan in a receding horizon within known free space $\envfree(t)$. In both cases, the planner would tend to guide the system into the narrow dead end, leading to a crash (recall that the planner's speed $v$ is fixed).

By contrast, our approach eventually takes a more circuitous---but recursively feasible---route to the goal. The evolution of planned viable trajectories is shown on the right in Fig.~\ref{fig:front}. Initially, we plan tight loops near $\home$, but over time we visit more of the safely explorable space $\explorable(\home)$, and eventually we find $\goal$.



%% file: conclusion.tex
\section{Discussion \& Conclusion}
\label{sec:conclusion}

In this paper, we have introduced a novel framework for recursively feasible motion planning for dynamical systems. Our approach is based on the ideas of forward and backward reachability, and uses FaSTrack \cite{herbert2017fastrack} to make a strong guarantee of safety over all time. Moreover, we also guarantee that if the initial ``home'' state is live, i.e. the goal is safely explorable from the home state, then each state along all motion plans will also be live, and eventually we will find a trajectory to the goal.

To our knowledge, this is the first motion planning algorithm to make this guarantee of recursive feasibility. As such, we have presented it as generally as possible and without optimization. While we make no claims of optimality, we do believe that many of the advances in optimal sample-based planning could be readily applied to our work. We are also sanguine about implementing our work in hardware for different, more complicated dynamical systems.